\documentclass[twoside,11pt]{article}

\usepackage{tikz}
\usetikzlibrary{positioning, arrows.meta, calc}

\usepackage[preprint]{jmlr2e}
\usepackage{xcolor}
\usepackage{listings}
\definecolor{kwblue}{rgb}{0.13,0.29,0.70}
\definecolor{strgreen}{rgb}{0.13,0.55,0.13}
\definecolor{commentgray}{rgb}{0.45,0.45,0.45}
\definecolor{backgray}{rgb}{0.97,0.97,0.97}
\usepackage{lastpage}
\jmlrheading{1}{2025}{1-\pageref{LastPage}}{}{}{}{Pariente and Indelman}

\ShortHeadings{POMDPPlanners}{Pariente and Indelman}
\firstpageno{1}

\begin{document}
	
	\title{POMDPPlanners: Open-Source Package for POMDP Planning}
	
	\author{\name Yaacov Pariente \email yaacovp@campus.technion.ac.il \\
		\addr Faculty of Mathematics\\
		Technion - Israel Institute of Technology\\
		Haifa 3200003, Israel
		\AND
		\name Vadim Indelman \email vadim.indelman@technion.ac.il \\
		\addr Stephen B. Klein Faculty of Aerospace Engineering\\
		Faculty of Data and Decision Sciences\\
		Technion - Israel Institute of Technology\\
		Haifa 3200003, Israel}

	\editor{}

	\maketitle
	
	\begin{abstract}
		We present POMDPPlanners, an open-source Python package for empirical evaluation of Partially Observable Markov Decision Process (POMDP) planning algorithms. The package integrates state-of-the-art planning algorithms, a suite of benchmark environments with safety-critical variants, automated hyperparameter optimization via Optuna, persistent caching with failure recovery, and configurable parallel simulation---reducing the overhead of extensive simulation studies. POMDPPlanners is designed to enable scalable, reproducible research on decision-making under uncertainty, with particular emphasis on risk-sensitive settings where standard toolkits fall short.
	\end{abstract}
	
	
	\section{Introduction}
	Partially Observable Markov Decision Processes (POMDPs) model sequential decision-making under uncertainty, with applications spanning robotics, autonomous navigation, and medical decision-making. Conducting robust, reproducible simulation studies, however, remains practically challenging: researchers must manually integrate planners, belief representations, and environments from disparate sources, implement hyperparameter search routines, and manage interrupted campaigns---all without standardized tooling in Python. Existing frameworks include C++ toolkits \citep{JMLR:ai_toolbox, JMLR:madp_toolbox} and the Julia-based POMDPs.jl \citep{JMLR:julia_pomdp}, while Python alternatives \citep{emami2015pomdpy, zheng2020pomdp_py} implement only basic solvers and do not support continuous spaces. We present POMDPPlanners, an open-source Python package that addresses these gaps with a unified, extensible framework for POMDP planning research. Table~\ref{tab:comparison} summarizes the key differences.

\begin{table}[t]
\centering
\small
\caption{Comparison of POMDP planning frameworks. \checkmark~= supported; -- = not supported. All frameworks support discrete spaces; Cont.\ = continuous spaces; Hyp.\ opt.\ = automated hyperparameter optimization; Safety Envs.\ = safety-critical environments with violation metrics; Modern Planners = includes planners beyond POMCP appearing in post-2018 POMDP literature (e.g., POMCPOW, PFT-DPW, BetaZero). $^\dagger$POMDPs.jl provides a solver interface; planners are separate community packages.}
\label{tab:comparison}
\resizebox{\textwidth}{!}{%
\begin{tabular}{lccccccc}
\hline
\textbf{Package} & \textbf{Language} & \textbf{Cont.} & \textbf{Modern Planners} & \textbf{Hyp. opt.} & \textbf{Safety Envs.} & \textbf{Parallel} & \textbf{Caching} \\
\hline
AI-Toolbox  & C++    & --         & --         & --         & --         & --         & --         \\
MADP        & C++    & --         & --         & --         & --         & --         & --         \\
POMDPs.jl   & Julia  & \checkmark & \checkmark$^\dagger$ & -- & --    & \checkmark & --         \\
pomdpy      & Python & --         & --         & --         & --         & --         & --         \\
pomdp\_py   & Python & --		  & --         & --         & --         & --         & --         \\
\textbf{POMDPPlanners} & \textbf{Python} & \checkmark & \checkmark & \checkmark & \checkmark & \checkmark & \checkmark \\
\hline
\end{tabular}}
\end{table}
	
	\section{Architecture}

	\begin{figure}[t]
	\begin{minipage}[c]{0.63\textwidth}
	\centering
	\begin{tikzpicture}[scale=0.62, transform shape,
	  wf/.style  = {rectangle, rounded corners=3pt, draw=black!70, thick,
	                fill=blue!12, minimum width=2.8cm, minimum height=0.85cm,
	                align=center, font=\small},
	  tm/.style  = {rectangle, rounded corners=3pt, draw=black!70, thick,
	                fill=orange!15, minimum width=11cm, minimum height=0.85cm,
	                align=center, font=\small},
	  ab/.style  = {rectangle, rounded corners=3pt, draw=black!70, thick,
	                fill=teal!12, minimum width=2.2cm, minimum height=0.85cm,
	                align=center, font=\small},
	  res/.style = {rectangle, rounded corners=3pt, draw=black!70, thick,
	                fill=green!12, minimum width=11cm, minimum height=0.85cm,
	                align=center, font=\small},
	  arr/.style = {-{Stealth[scale=1.0]}, thick},
	  int/.style = {-{Stealth[scale=0.85]}, dashed, semithick},
	]
	\node[wf] (wf1) at (0.5,  0)    {Direct\\Evaluation};
	\node[wf] (wf2) at (4.5,  0)    {Optimize-and-Evaluate\\(\textit{Optuna})};
	\node[tm] (tm)  at (2.5, -1.6)  {\textbf{Task Manager} \& Persistent Cache};
	\node[ab] (env)    at (-1.5, -3.4) {\texttt{Environment}};
	\node[ab] (belief) at (2.5,  -3.4) {\texttt{Belief}};
	\node[ab] (policy) at (6.5,  -3.4) {\texttt{Policy}};
	\node[res] (out) at (2.5, -5.0)  {Results \& MLflow Logging};
	\coordinate (jtm)  at (2.5, -0.8);
	\coordinate (jtb)  at (2.5, -2.4);
	\coordinate (jout) at (2.5, -4.2);
	\draw      (wf1.south) |- (jtm);
	\draw      (wf2.south) |- (jtm);
	\draw[arr] (jtm) -- (tm.north);
	\draw      (tm.south) -- (jtb);
	\draw[arr] (jtb) -| (env.north);
	\draw[arr] (jtb) -- (belief.north);
	\draw[arr] (jtb) -| (policy.north);
	\draw[int] (env.east)    -- node[above, font=\scriptsize\itshape] {dynamics} (belief.west);
	\draw[int] (belief.east) -- (policy.west);
	\draw      (env.south)    |- (jout);
	\draw      (belief.south) -- (jout);
	\draw      (policy.south) |- (jout);
	\draw[arr] (jout) -- (out.north);
	\end{tikzpicture}
	\end{minipage}%
	\hfill
	\begin{minipage}[c]{0.34\textwidth}
	\caption{Architecture of \texttt{POMDPPlanners}. Workflows feed into a shared Task Manager with persistent caching. Dashed arrows indicate data flow between abstractions; outputs are logged via MLflow.}
	\label{fig:architecture}
	\end{minipage}
	\end{figure}

	\subsection{Core Abstractions}
	\texttt{Environment} defines POMDP dynamics via state transition and observation models, reward function, terminal condition, and initial distributions; a \texttt{SpaceInfo} dataclass specifies discrete, continuous, or mixed spaces, enabling runtime compatibility checking. \texttt{Belief} maintains a distribution over states and performs Bayesian updates using the environment's models; supported representations include weighted and unweighted particle filters, Gaussian, and Gaussian mixture beliefs. \texttt{Policy} exposes an \texttt{action(belief)} method returning an action together with a \texttt{PolicyRunData} structure of per-step diagnostics (e.g., expanded nodes, visit counts in MCTS-based planners), supporting post-hoc analysis of planner behavior.

	\subsection{Workflows and Experiment Management}
	POMDPPlanners supports two complementary workflows: \textit{direct evaluation}, which returns aggregated statistics (mean return, Conditional Value-at-Risk (CVaR), Value-at-Risk (VaR), confidence intervals), and \textit{optimize-and-evaluate}, which runs an Optuna \citep{akiba2019optuna} search over each planner's hyperparameter space before forwarding the best configuration to evaluation. Both workflows share a fault-tolerant task manager: each simulation is keyed by a SHA-256 hash of its full specification (environment parameters, policy, belief, seed), so cache hits return instantly and interrupted experiments resume automatically. The task manager is backed by a pluggable execution layer supporting Joblib (multi-core), Dask (distributed, including multi-machine clusters via a scheduler address), and PBS (HPC batch queues via \texttt{dask-jobqueue}); switching backends requires only changing a single configuration object. Per-episode returns, safety metrics, hyperparameters, and \texttt{PolicyRunData} diagnostics are recorded as MLflow runs, enabling cross-experiment comparison via the MLflow UI or programmatic queries. The framework optionally caches a GIF of each episode's trajectory alongside numerical results, since aggregate statistics can be misleading---e.g., a zero obstacle-hit rate may simply reflect an agent that never moved. The package exposes flexible interfaces for custom planners, environments, and metrics.

	\section{Benchmark Environments and Algorithms}
	\label{sec:envs}

	\subsection{Benchmark Environments}
	The package includes nine benchmark environments: Tiger, LightDark, RockSample, CartPole, MountainCar, Push, LaserTag, SafetyAnt, and PacMan. LightDark and LaserTag each provide discrete and continuous variants. Most environments incorporate configurable dangerous areas with associated penalty rewards; the specific mechanisms vary: LightDark exposes stochastic obstacle hits via \texttt{obstacle\_hit\_probability} and an \texttt{is\_obstacle\_hit\_terminal} flag; LaserTag and RockSample provide configurable \texttt{dangerous\_areas} with tunable penalty magnitudes; SafetyAnt enforces a hard velocity safety constraint with a critical-violation termination threshold. Risk-enriched environments report dedicated safety metrics (violation rates, total counts) alongside standard return statistics.

	\subsection{Planning Algorithms}
	The package provides the following online planners: POMCP \citep{silver2010monte}, POMCP-DPW, POMCPOW \citep{sunberg2018online}, SparsePFT, PFT-DPW \citep{sunberg2018online}, Sparse Sampling \citep{Kearns02jml}, DiscreteActionSequences (open-loop baseline), BetaZero \citep{moss2024betazero}, and the following risk-averse planners for safe planning: ConstrainedZero \citep{moss2024constrainedzero},  ICVaR Sparse Sampling, ICVaR POMCPOW, and ICVaR PFT-DPW \citep{pariente2026icvar}. Each planner exposes a standardized hyperparameter interface compatible with Optuna search. The selection covers state-of-the-art planners commonly used as baselines in recent POMDP literature.
	
	\section{Usage Example}
	The listing illustrates direct evaluation of POMCPOW and PFT-DPW on \texttt{ContinuousLightDarkPOMDP\-DiscreteActions}. Two planners and a vectorized particle-filter belief are passed to \texttt{LocalSimulationsAPI}, which runs parallelized simulations and returns aggregated statistics. For hyperparameter optimization, \texttt{run\_optimize\_and\_evaluate()} accepts \texttt{HyperParameterRunParams} with Optuna search ranges and forwards the best configuration to evaluation automatically.

\begin{lstlisting}
from POMDPPlanners.environments import ContinuousLightDarkPOMDPDiscreteActions
from POMDPPlanners.planners.mcts_planners.pomcpow import POMCPOW
from POMDPPlanners.planners.mcts_planners.pft_dpw import PFT_DPW
from POMDPPlanners.utils.action_samplers import DiscreteActionSampler
from POMDPPlanners.utils.belief_factory import create_environment_belief
from POMDPPlanners.simulations.simulation_apis.local_simulations_api import (
    LocalSimulationsAPI)
from POMDPPlanners.core.simulation import EnvironmentRunParams

env    = ContinuousLightDarkPOMDPDiscreteActions(discount_factor=0.95)
sampler = DiscreteActionSampler(env.get_actions())
pomcpow = POMCPOW(environment=env, discount_factor=0.95, depth=10,
                  exploration_constant=10.0, k_o=2.0, k_a=2.0,
                  alpha_o=0.5, alpha_a=0.5, n_simulations=500,
                  action_sampler=sampler, name="POMCPOW")
pft_dpw = PFT_DPW(environment=env, discount_factor=0.95, depth=10,
                  exploration_constant=10.0, n_simulations=500,
                  action_sampler=sampler, name="PFT_DPW")
belief  = create_environment_belief(env, n_particles=200)

api = LocalSimulationsAPI()
_, stats = api.run_multiple_environments_and_policies(
    environment_run_params=[EnvironmentRunParams(
        environment=env, belief=belief,
        policies=[pomcpow, pft_dpw], num_episodes=100, num_steps=30)],
    alpha=0.1, experiment_name="LightDark_Evaluation")
\end{lstlisting}

	The call returns aggregated statistics including mean return, CVaR, goal rate, and safety metrics, alongside MLflow-logged per-episode data.

	\paragraph{Conclusions.} We have presented POMDPPlanners, an open-source Python framework that unifies state-of-the-art POMDP planners, risk-enriched benchmark environments, and automated experimentation tooling, lowering the barrier to rigorous simulation studies so that researchers can focus on algorithmic contributions rather than infrastructure.

	\paragraph{Availability.} Source code (MIT license), documentation, and Jupyter notebook examples are available at \url{https://github.com/yaacovpariente/POMDPPlanners}. The package requires Python~3.10+ and is installed via \texttt{pip install POMDPPlanners}.
	\vskip 0.1in
	\bibliography{references_open_source.bib}

@misc{pariente2026icvar,
    author    = {Pariente, Yaacov and Indelman, Vadim},
    title     = {Online Risk-Averse Planning in {POMDP}s Using Iterated {CVaR} Value Function},
    year      = {2026},
    eprint    = {2601.20554},
    archivePrefix = {arXiv},
    primaryClass  = {cs.AI},
    doi       = {10.48550/arXiv.2601.20554}
}

@article{Kearns02jml,
    author =     {Kearns, Michael and Mansour, Yishay and Ng, Andrew Y},
    journal =     {Machine learning},
    number =     2,
    pages =     {193--208},
    publisher =     {Springer},
    title =     {A sparse sampling algorithm for near-optimal planning in large Markov decision processes},
    volume =     49,
    year =     2002
}

@article{JMLR:ai_toolbox,
	author  = {Eugenio Bargiacchi and Diederik M. Roijers and Ann NowÃ©},
	title   = {AI-Toolbox: A C++ library for Reinforcement Learning and Planning (with Python Bindings)},
	journal = {Journal of Machine Learning Research},
	year    = {2020},
	volume  = {21},
	number  = {102},
	pages   = {1--12},
	url     = {http://jmlr.org/papers/v21/18-402.html}
}

@article{JMLR:madp_toolbox,
	author  = {Frans A. Oliehoek and Matthijs T. J. Spaan and Bas Terwijn and Philipp Robbel and JoÃ£o V. Messias},
	title   = {The MADP Toolbox: An Open Source Library for Planning and Learning in (Multi-)Agent Systems},
	journal = {Journal of Machine Learning Research},
	year    = {2017},
	volume  = {18},
	number  = {89},
	pages   = {1--5},
	url     = {http://jmlr.org/papers/v18/17-156.html}
}

@inproceedings{zheng2020pomdp_py,
   title = {pomdp\_py: A Framework to Build and Solve POMDP Problems},
   author = {Zheng, Kaiyu and Tellex, Stefanie},
   booktitle = {ICAPS 2020 Workshop on Planning and Robotics (PlanRob)},
   year = {2020},
   url = {https://icaps20subpages.icaps-conference.org/wp-content/uploads/2020/10/14-PlanRob_2020_paper_3.pdf},
   note = {Arxiv link: "\url{https://arxiv.org/pdf/2004.10099.pdf}"}
}

@article{JMLR:julia_pomdp,
  author  = {Maxim Egorov and Zachary N. Sunberg and Edward Balaban and Tim A. Wheeler and Jayesh K. Gupta and Mykel J. Kochenderfer},
  title   = {POMDPs.jl: A Framework for Sequential Decision Making under Uncertainty},
  journal = {Journal of Machine Learning Research},
  year    = {2017},
  volume  = {18},
  number  = {26},
  pages   = {1--5},
  url     = {http://jmlr.org/papers/v18/16-300.html}
}

@article{silver2010monte,
	title={Monte-Carlo planning in large POMDPs},
	author={Silver, David and Veness, Joel},
	journal={Advances in neural information processing systems},
	volume={23},
	year={2010}
}

@inproceedings{sunberg2018online,
	title={Online algorithms for POMDPs with continuous state, action, and observation spaces},
	author={Sunberg, Zachary and Kochenderfer, Mykel},
	booktitle={Proceedings of the International Conference on Automated Planning and Scheduling},
	volume={28},
	pages={259--263},
	year={2018}
}

@inproceedings{moss2024betazero,
	title={{BetaZero: Belief-State Planning for Long-Horizon POMDPs using Learned Approximations}},
	author={Moss, Robert J. and Corso, Anthony and Caers, Jef and Kochenderfer, Mykel J.},
	booktitle={Reinforcement Learning Conference (RLC)},
	year={2024},
}

@inproceedings{moss2024constrainedzero,
	title={{ConstrainedZero: Chance-Constrained POMDP Planning Using Learned Probabilistic Failure Surrogates and Adaptive Safety Constraints}},
	author={Moss, Robert J. and Jamgochian, Arec and Fischer, Johannes and Corso, Anthony and Kochenderfer, Mykel J.},
	booktitle={International Joint Conference on Artificial Intelligence (IJCAI)},
	year={2024},
}

@inproceedings{akiba2019optuna,
  title={{O}ptuna: A Next-Generation Hyperparameter Optimization Framework},
  author={Akiba, Takuya and Sano, Shotaro and Yanase, Toshihiko and Ohta, Takeru and Koyama, Masanori},
  booktitle={The 25th ACM SIGKDD International Conference on Knowledge Discovery \& Data Mining},
  pages={2623--2631},
  year={2019}
}

@misc{emami2015pomdpy,
	author = {Emami, Patrick and Hamlet, Alan J. and Crane, Carl},
	title = {POMDPy: An Extensible Framework for Implementing POMDPs in Python},
	year = {2015},
}
	
\end{document}